\pdfoutput=1

\documentclass[11pt]{article}

\usepackage{acl}

\usepackage{times}
\usepackage{latexsym}

\usepackage{devanagari}
\usepackage{graphicx}
\usepackage[table]{colortbl}
\usepackage{booktabs}

\usepackage[T1]{fontenc}

\usepackage[utf8]{inputenc}

\usepackage{microtype}

\title{Gender Inflected or Bias Inflicted: On Using Grammatical Gender Cues for Bias Evaluation in Machine Translation}

\author{Pushpdeep Singh \\
  National Institute of Technology, Hamirpur \\ Anu, Hamirpur, India \\
\texttt{pushpdeep30@gmail.com} \\}

\begin{document}
\maketitle
\begin{abstract}
Neural Machine Translation (NMT) models are state-of-the-art for machine translation. However, these models are known to have various social biases, especially gender bias. Most of the work on evaluating gender bias in NMT has focused primarily on English as the source language. For source languages different from English, most of the studies use gender-neutral sentences to evaluate gender bias. However, practically, many sentences that we encounter do have gender information. Therefore, it makes more sense to evaluate for bias using such sentences. This allows us to determine if NMT models can identify the correct gender based on the grammatical gender cues in the source sentence rather than relying on biased correlations with, say, occupation terms. To demonstrate our point, in this work, we use Hindi as the source language and construct two sets of gender-specific sentences: \textit{OTSC-Hindi} and \textit{WinoMT-Hindi} that we use to evaluate different Hindi-English (HI-EN) NMT systems automatically for gender bias. Our work highlights the importance of considering the nature of language when designing such extrinsic bias evaluation datasets.

\end{abstract}

\section{Introduction}

Various models trained to learn from data are susceptible to picking up spurious correlations in their training data, which can lead to multiple social biases. In NLP, such biases have been observed in different forms: \citet{Bolukbasi2016ManIT} found that word embeddings exhibit gender stereotypes, \citet{zhao-etal-2017-men} observed that models for visual semantic role labelling aggrandize existing gender bias present in data, similar biased behaviour had been observed in NLP tasks like coreference resolution \citep{lu2019gender} and Natural Language Inference \citep{rudinger-etal-2017-social}.     

Even state-of-the-art NMT models develop such biases \citep{prates2019assessing}. These models can express gender bias in different ways. One is when due to their poor coreference resolution ability, they rely on biased associations with, say, occupation terms to disambiguate the gender of pronouns (\citealp{stanovsky-etal-2019-evaluating, saunders-etal-2020-neural}). Another is when these models translate gender-neutral sentences into gendered ones (\citealp{prates2019assessing, cho-etal-2019-measuring}). In many cases, NMT models give a `masculine default' translation. 

This problem also exists for HI-EN Machine Translation \citep{ramesh-etal-2021-evaluating}. When put to use, such systems can cause various harms \citep{savoldi-etal-2021-gender}. Thus, evaluating and mitigating such biases from NMT models is critical to ensure fairness. 

Prior research evaluating gender bias in machine translation has predominantly centered around English as the source language \citep{stanovsky-etal-2019-evaluating}. However, these evaluation methods or benchmarks don't seamlessly extend to other source languages, especially the ones with grammatical gender. For instance, in Hindi, elements like pronouns, adjectives, and verbs are often inflected with gender. Nonetheless, prior studies in other source languages often utilize gender-neutral sentences (\citealp{cho-etal-2019-measuring, ramesh-etal-2021-evaluating}) for bias evaluation. Yet, in practice, many sentences inherently possess gender information. 

Therefore, in this work, we propose to evaluate NMT models for bias using sentences with grammatical gender cues of the source language. This allows us to ascertain whether NMT models can discern the accurate gender from context or if they depend on biased correlations. In this work, we contribute the following :

\begin{itemize}
    \item Using Hindi as source language in NMT, we highlight the limitations of existing bias evaluation methods that use gender-neutral sentences.
    \item Additionally, we propose to use context-based gender bias evaluation using grammatical gender markers of the source language. We construct two evaluation sets for bias evaluation of NMT models: Occupation Testset with Simple Context (\textit{OTSC-Hindi}) and \textit{WinoMT-Hindi}. 
    \item Using these evaluation sets, we evaluate various blackbox and open-source HI-EN NMT models for gender bias.
    \item We highlight the importance of creating such benchmarks for source languages with expressive gender markers.
\end{itemize}
Code and data are publicly available\footnote{\url{https://github.com/iampushpdeep/Gender-Bias-Hi-En-Eval}}.

\section{Experimental Setup}
\textbf{NMT Models : }We test HI-EN NMT models which are widely popular and represent state-of-the-art in both commercial or academic research : (1) IndicTrans \citep{ramesh-etal-2022-samanantar}, (2) Google Translate\footnote{\url{https://translate.google.com/}}, (3) Microsoft Translator\footnote{\url{https://www.bing.com/translator}}, and (4) AWS Translate\footnote{\url{https://aws.amazon.com/translate/}}. IndicTrans is an academic, open-source multilingual NMT model, while the latter four are commercial NMT systems available via APIs.
\\ \\
\textbf{Hindi as Source Language : }We create bias evaluation sentences in Hindi to evaluate HI-EN NMT Models. We choose Hindi due to two reasons. First, only limited research has been done on evaluating gender bias in Hindi translation. Previous work by \citet{ramesh-etal-2021-evaluating} focused only on the gender-neutral side of Hindi by evaluating simple sentences with gender-neutral, third person pronouns like ``{\dn vh}(vah)'', ``{\dn v\?}(ve)'' and ``{\dn vo}(vo)''. Second, choosing Hindi allows us to demonstrate bias evaluation using sentences with a diverse range of gender markers. In Hindi, verbs, adjectives and possessive pronouns often carry gender indicators. The grammatical gender system in Hindi is exclusively rooted in biological gender \citep{agnihotri2007hindi}. However, the variety of gender markers can be different for different languages. Therefore it's essential to study gender-related rules of the specific language for creating benchmarks for such tasks. 

\section{TGBI Evaluation using Gender-Neutral Sentences}

\citet{cho-etal-2019-measuring} introduced \textit{translation gender bias index} (TGBI) as a metric to measure bias in NMT systems using gender-neutral source language sentences, originally for the Korean language. \citet{ramesh-etal-2021-evaluating} showed that the TGBI metric can be applied to Hindi too. They constructed seven sets ($P_1$ to $P_7$) of gender-neutral sentences in Hindi which included: formal (S1), impolite (S2), informal (S3), occupation (S4), negative (S5), polite (S6), and positive (S7) versions. 

For translation into English, TGBI uses the
fraction of sentences in a sentence set $S$ translated as ``masculine'', ``feminine'' or ``neutral" in the target , i.e., $p_{m}$, $p_{f}$ and $p_{n}$, respectively to calculate $P_S$ as :
        \begin{equation}
            P_{S} = \sqrt{(p_{m}p_{f}+p_{n})}
        \end{equation}
$P_i$ is calculated for each sentence set $S_i$ ($S_1$ to $S_n$) to finally calculate TGBI = avg($P_{i}$). Using lists from \citet{ramesh-etal-2021-evaluating}, we evaluate four HI-EN NMT models using the TGBI score to create a comparison for our evaluation methods. 

Often, using a metric like TGBI is not very practical. For example, when the original intent is not gender-neutral but constraints of the source language make it gender-neutral, then showing all versions\footnote{\url{https://ai.googleblog.com/2020/04/a-scalable-approach-to-reducing-gender.html}} or  \textit{random guessing}, with a 50\% chance of choosing one gender in translation, are more practical. Also, gender-specific sentences are more common and making errors in such sentences makes for a more unfair system. Hence, we propose to expose gender bias by evaluating NMT models on such source language sentences.

\section{Approach}

We construct two sets of sentences, one with a simple gender-specified context and another with a more complex context. In creating these sets, we focus on the gender markers of the source language, i.e. Hindi. Also, we use template sentences which can help to automatically evaluate bias without using additional tools at the target side.

\begin{figure}[tbh!]
\centering
\includegraphics[width=\columnwidth]{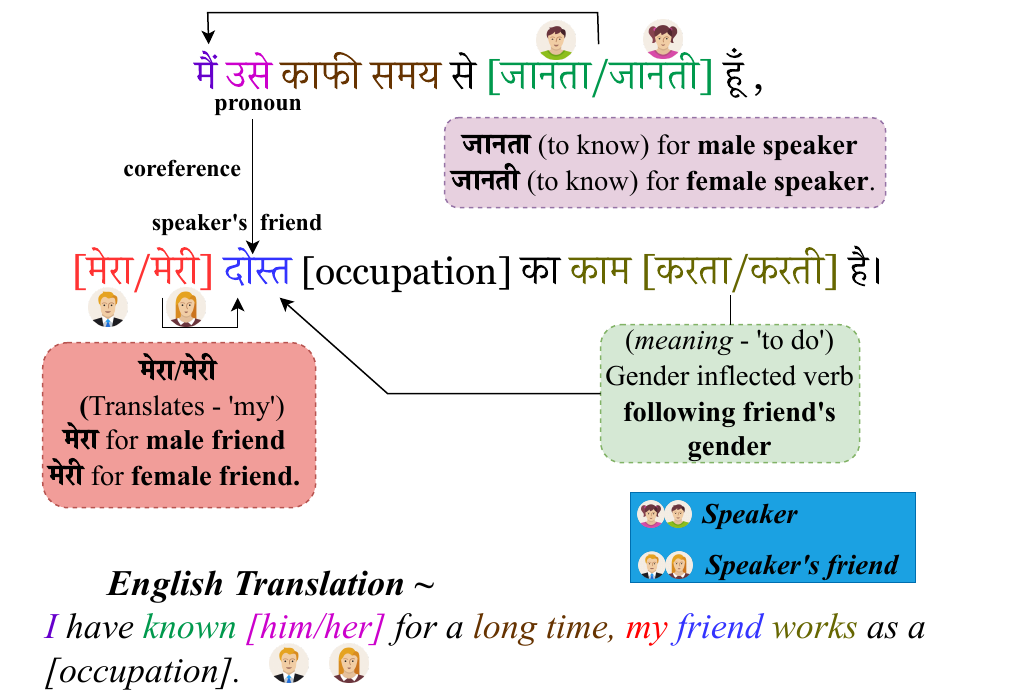}
\caption{OTSC-Hindi sample template sentence along with its English translation. Gender of the speaker is specified by gender-inflected verb, i.e. ``{\dn jAntA}'' or ``{\dn jAntF}''. The possessive pronoun ``{\dn m\?rA}" or ``{\dn m\?rF}" and the verb ``{\dn krtA}'' or ``{\dn krtF}'' specify friend's gender. Here, the pronoun ``{\dn us\?}" references speaker's friend.}
\label{fig:simplecontext}
\end{figure}

\subsection{OTSC-Hindi}

\citet{escude-font-costa-jussa-2019-equalizing} created a test set with custom template sentences to evaluate the gender bias for English to Spanish Translation. Inspired by this template, we create a Hindi version with grammatical gender cues: `` {\dn m\4{\qva} us\? kAPF smy s\?} \{{\dn jAntA{\rs ,\re}jAntF}\} {\dn \8{h}\1{\rs,\re}}\{{\dn m\?rA{\rs,\re}m\?rF}\} {\dn do-t} \textbf{[occupation]} {\dn kA kAm} \{{\dn krtA{\rs ,\re}krtF}\} {\dn h\4.} '' (\textit{I have known [him/her] for a long time, my friend works as a [occupation].}) 
Figure~\ref{fig:simplecontext} explains the template and gender-related information. Note that, unlike the English version, this template specifies the gender of the speaker (first person) using a gender-inflected verb, i.e. ``{\dn jAntA}(\textit{jaanta})'' for male while ``{\dn jAntF}(\textit{jaanti})'' for female. The possessive pronoun is also gender inflected based on the gender of the speaker's friend. In Hindi, the possessive pronoun is gender inflected based on the word following it, here ``{\dn m\?rA}(\textit{mera})" is used for male friend while ``{\dn m\?rF}(\textit{meri})" is used for female friend. Based on the use of ``{\dn m\?rA}(\textit{mera})" or ``{\dn m\?rF}(\textit{meri})", the verb ``{\dn krtA}(\textit{karta})'' and ``{\dn krtF}(\textit{karti})'' is used for a male friend and female friend, respectively. So in this template, there are four possibilities based on the gender of the speaker and the gender of the speaker's friend. Using 1071 occupations, we construct these four sets with 1071 sentences each and check the percentage of sentences where the speaker's friend is translated as male or female. This is because English translation only specifies the gender of the friend while the gender of the speaker is lost in translation.

\begin{figure}[tbh!]
\centering
\includegraphics[width=\columnwidth]{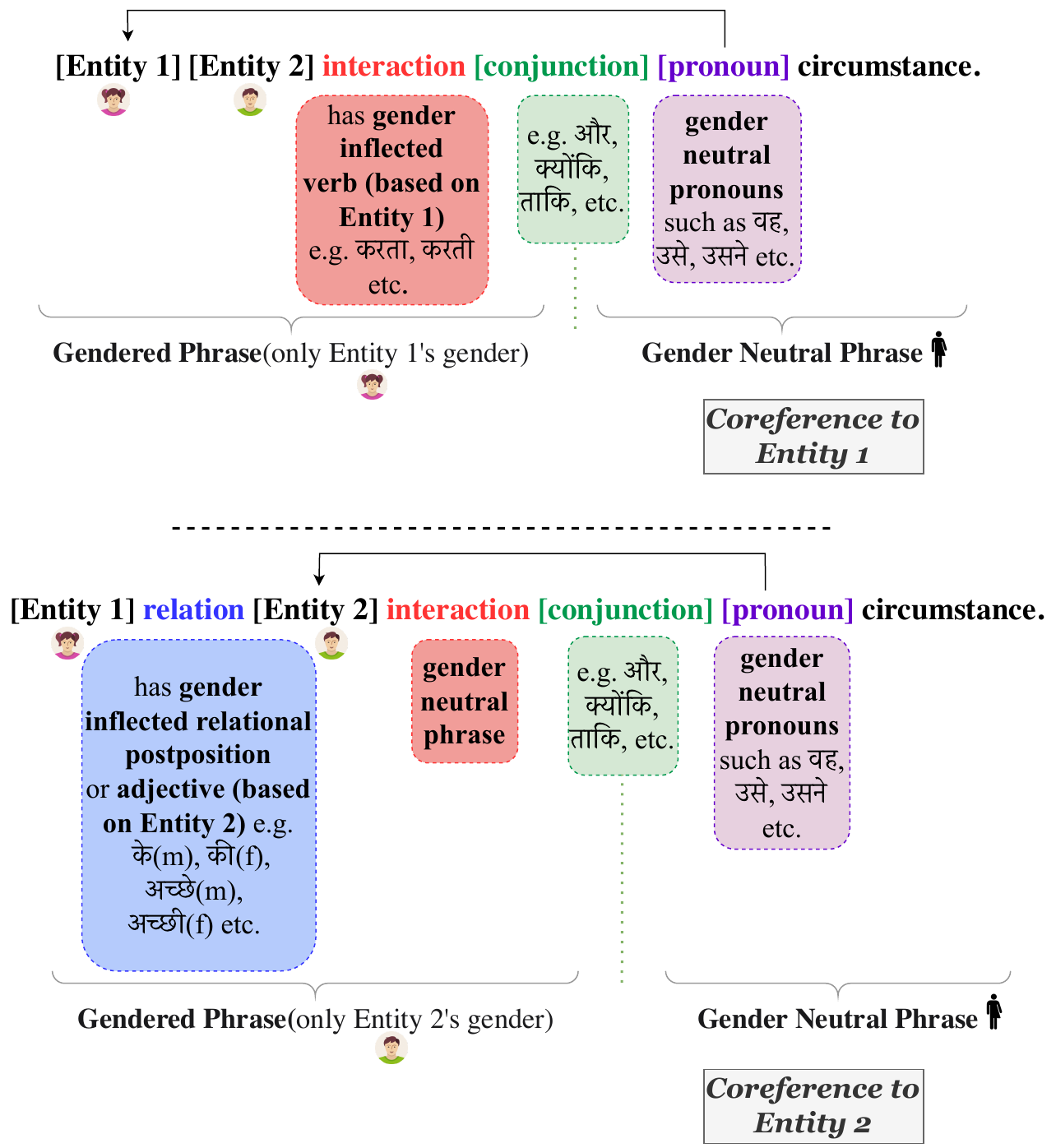}
\caption{Sentence Template for WinoMT-Hindi. When Entity 1 is referenced, we use gender-inflected verb to specify its gender. When Entity 2 is referenced, its gender is specified using gender-inflected relational postposition or an adjective. Phrase after the conjuction (containing the pronoun which refers to either entity) is gender neutral.}
\label{fig:winomthi}
\end{figure}

\begin{figure}[htb!]
\centering
\includegraphics[width=\columnwidth]{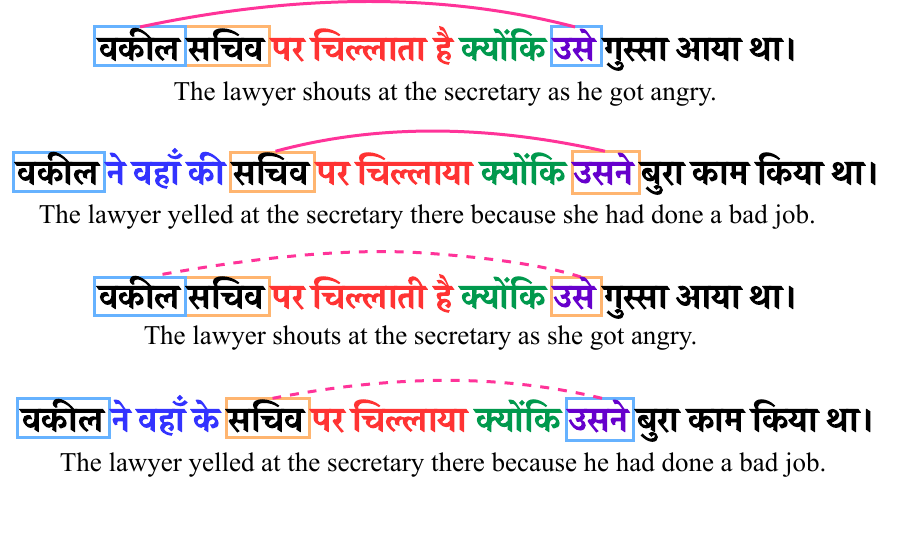}
\caption{Sample Sentences in WinoMT-Hindi. The solid line shows pro-stereotypical coreference, while the dashed line shows anti-stereotypical coreference. Male and female (stereotypically) entities are marked in blue and orange boxes, respectively. Hindi pronouns are marked in blue or orange box based on the actual gender of their referred entity according to the grammatical context.}
\label{fig:samplewinomthi}
\end{figure}

\subsection{WinoMT-Hindi}

In the real world, NMT models deal with more complex sentences: long sentences with further context, more entities, and complex coreferences. A model not good at coreference resolution becomes gender-biased when it relies on biased correlations to ascertain gender. \citet{stanovsky-etal-2019-evaluating} composed a challenge set called \textit{WinoMT} for evaluating gender bias in NMT models. This set combined gender-balanced \textit{Winogender} \citep{rudinger-etal-2018-gender} and \textit{WinoBias} \cite{zhao-etal-2018-gender} coreference test sets. However, since it is in English, using it for evaluating bias for other source languages is not possible. Therefore we contextualize this test set for the evaluation of bias in HI-EN Translation by manually creating ``WinoMT-Hindi'', which consists of 704 WinoBias-like sentences in Hindi, but modified to include gender cues of the language, mainly: gender-inflected adjectives, postpositions, and verbs. 

Construction of ``WinoMT-Hindi'' is explained in Figure~\ref{fig:winomthi}. Sample sentences are shown in Figure~\ref{fig:samplewinomthi}. In Winobias, only the English pronoun carries the gender of referenced entity, but here, to provide the gender of the referenced entity, we use gender-inflected verbs for Entity 1 and postpositions or adjectives for Entity 2. The phrase after the conjunction is gender-neutral, challenging the model to look for a more extended context. We only specify the gender of the referenced entity to avoid confusing the model with too much information. 

We don't need reference translations in English, as automatic evaluation is possible. Due to the nature of our source sentences, we can mark the gender of the target by simply checking for the presence of male pronouns (he, him or his) or female pronouns (she or her) in the translation. 
Interestingly, we also observe that few sentences are translated into gender-neutral form. For example, the sentence: ``{\dn sEcv \8{m}vr s\? \8{p}CtA h\4 Ek vh mdd k\? Ele \3C8wA kr\?}'' (Secretary asks mover what he should do to help) is translated as ``The secretary asks the mover what to do to help'' by Google Translate. While there is an increased interest in promoting Gender-Neutral translation for inclusivity \citep{piergentili2023gender}, others call for gender preservation in translation \citep{cabrera2023gender}. The presence of neutral output sentences can be modelled as \textit{false negatives} or \textit{true positives} depending upon the goals of the evaluation. For this study, we model their presence as \textit{false negatives} for male and female class, i.e. equivalent to misgendering sentences. Nonetheless, due to the limited fraction of such sentences, metrics largely reflect bias due to misgendering.

For gender bias evaluation, we use the metrics: \emph{Acc}, $\Delta_G$ and $\Delta_S$ given by \citet{stanovsky-etal-2019-evaluating}. For measuring the difference in $F_1$ score between male and female classes, i.e. $\Delta_G$, we use class-wise $F_1$ score. We have divided our sentences into pro-stereotypical and anti-stereotypical sets using translated and transliterated versions of the occupations list by \citet{zhao-etal-2018-gender}. This was done manually to ensure gender-neutrality of these occupation terms (and avoid their gender-inflected versions) in Hindi. To measure the difference in overall performance between pro-stereotypical and anti-stereotypical groups, i.e., $\Delta_S$, we use $macro$-$F_1$ score by averaging $F_1$ for male and female class only. We also report the percentage of sentences translated as gender-neutral, i.e. \emph{N} for each NMT system.

\begin{table}[tbh!]
  \centering
  \begin{tabular}{@{}lllll@{}}
    \toprule
    & \multicolumn{1}{c}{\small{\textbf{IT}}} & \multicolumn{1}{c}{\small{\textbf{GT}}} & \multicolumn{1}{c}{\small{\textbf{MS}}} & \multicolumn{1}{c}{\small{\textbf{AWS}}} \\ \midrule
    \textit{\small{S1}} &0.787 &0.708 &\textbf{0.724} &0.691    \\
    \textit{\small{S2}} &0.620 &0.534 &0.394 &0.656     \\ 
    \textit{\small{S3}} &0.623 &0.623 &0.467 &0.682     \\ 
    \textit{\small{S4}} &0.569 &0.531 &0.574 &0.411     \\ 
    \textit{\small{S5}} &\textbf{0.819} &\textbf{0.763} &0.673 &\textbf{0.803}     \\ 
    \textit{\small{S6}} &\textbf{0.926$^{*}$} &\textbf{0.862$^{*}$} &\textbf{0.951$^{*}$} &\textbf{0.725}     \\ 
    \textit{\small{S7}} &\textbf{0.848} &\textbf{0.788} &\textbf{0.720} &\textbf{0.845$^{*}$}     \\ 
    \midrule
    \textit{\textbf{TGBI}} &\cellcolor{black!25}0.742 &0.687 &0.643 &0.688 \\
    \bottomrule
    \end{tabular}%
  \caption{TGBI Evaluation of IndicTrans (IT), Google Translate (GT), Microsoft Translator (MS) and AWS Translate (AWS). The table contains the $P$ values (higher is better) and their average, i.e. TGBI at the bottom. Bold represents the top three highest $P$ values. $^{*}$ represent set with highest $P$ value. The highlighted cell represents the highest TGBI value.}
  \label{tab:results-tgbi}
  \end{table}
\begin{table*}[tbh!]
  \centering
  \begin{tabular}{@{}lcccccccc@{}}
    \toprule
    & \multicolumn{2}{c}{\textbf{IT}}                                                                & \multicolumn{2}{c}{\textbf{GT}}                                                                & \multicolumn{2}{c}{\textbf{MS}}                                                                 & \multicolumn{2}{c}{\textbf{AWS}}                                                             \\
\multicolumn{1}{c}{\textbf{Sentence Set}}    & \multicolumn{1}{c}{$p_m$} & \multicolumn{1}{c}{$p_w$} & \multicolumn{1}{c}{$p_m$} & \multicolumn{1}{c}{$p_w$} & \multicolumn{1}{c}{$p_m$} & \multicolumn{1}{c}{$p_w$}  & \multicolumn{1}{c}{$p_m$} & \multicolumn{1}{c}{$p_w$} \\ \midrule
    \multicolumn{1}{l|}{\textit{\small{Female Speaker, Female Friend}}} & \textbf{98.41}                                                  & \multicolumn{1}{c|}{1.59$^*$}      & 1.68                         & \multicolumn{1}{c|}{\textbf{98.32$^*$}}      & \textbf{98.97}                                   & \multicolumn{1}{c|}{1.03$^*$}      & \textbf{95.61}                                                    & 4.39$^*$                           \\
    \multicolumn{1}{l|}{\textit{\small{Female Speaker, Male Friend}}} & {\textbf{99.25$^*$}}                                  & \multicolumn{1}{c|}{0.75}      & \textbf{90.66$^*$}  & \multicolumn{1}{c|}{9.34}      & \textbf{99.72$^*$}                                                     & \multicolumn{1}{c|}{0.28}      & \textbf{95.70$^*$}                                              & 4.30                            \\
    \multicolumn{1}{l|}{\textit{\small{Male Speaker, Female Friend}}} & \textbf{99.35}                                                   & \multicolumn{1}{c|}{0.65$^*$}      & 2.43                                & \multicolumn{1}{c|}{\textbf{97.57$^*$}}      & \textbf{66.01}                                       & \multicolumn{1}{c|}{33.99$^*$}      & \textbf{99.29}                                                    & 2.71$^*$                            \\
        \multicolumn{1}{l|}{\textit{\small{Male Speaker, Male Friend}}} & \textbf{99.91$^*$}                                                      & \multicolumn{1}{c|}{0.09}      & \textbf{96.45$^*$}                               & \multicolumn{1}{c|}{3.55}      & \textbf{98.60$^*$}                                          & \multicolumn{1}{c|}{1.40}      & \textbf{97.48$^*$}                                               & 2.52                            \\\bottomrule
    \end{tabular}%

  \caption{
    Evaluation of IndicTrans(IT), Google Translate(GT), Microsoft Translator(MS) and AWS Translate(AWS) using the OTSC-Hindi test set. Here $p_m$ and $p_w$ are the percentage of sentences translated as male and female, respectively for the speaker’s friend. $*$ corresponds to the percentage of sentences translated into the true label for each sentence set. Bold values indicate the maximum percentage of sentences translated into a single gender class.}
    \label{tab:results_otsc}
\end{table*}
\begin{table}[tbh!]
  \centering
  \begin{tabular}{@{}lcccc@{}}
    \toprule
    & \multicolumn{1}{c}{\textbf{$Acc$}} & \multicolumn{1}{c}{\textbf{$\Delta_G$}} & \multicolumn{1}{c}{\textbf{$\Delta_S$}} & \multicolumn{1}{c}{\textbf{$N$}} \\ \midrule
    \textit{\small{\textbf{IndicTrans}}} &48.9 &48.5 &-0.1$^{\bullet}$ &6.2    \\
    \textit{\small{\textbf{Google Translate}}} &69.0$^{\star}$ &10.6$^{\diamond}$ &-3.8 &5.3     \\ 
    \textit{\small{\textbf{Microsoft Translator}}} &57.7 &32.9 &0.2$^{\bullet}$ &4.1     \\ 
    \textit{\small{\textbf{AWS Translate}}} &49.9 &51.9 &-0.2$^{\bullet}$ &2.8     \\ 
    \bottomrule
    \end{tabular}%
  \caption{Comparison of performance of various NMT Models on WinoMT-Hindi on \emph Acc, $\Delta_G$, $\Delta_S$ and $N$ (all in \%) measures. ${\star}$ indicates significantly highest value, ${\diamond}$ indicates significantly lowest value, ${\bullet}$ indicates near about values for \emph Acc, $\Delta_G$ and $\Delta_S$, respectively. }
  \label{tab:results-winomt}
  \end{table}

\section{Results and Discussion}
    \subsection{TGBI Evaluation}

        The results are shown in Table~\ref{tab:results-tgbi}. For most translation systems, sentences in ``Negative (S5)'', ``Polite (S6)'' and ``Positive (S7)'' sets have higher $P$ values. With the highest TGBI score, ``IndicTrans'' performs better at translating gender-neutral Hindi sentences into English with minimum gender bias. The problem with the TGBI metric is that it may not accurately capture the true fairness of an NMT system since evaluation is only done on gender-neutral sentences. 
        
    \subsection{Evaluation using OTSC-Hindi}

        The results are shown in Table~\ref{tab:results_otsc}. Based on these results, the IndicTrans system shows heavy bias against the feminine gender. Even though it has the highest TGBI score, IndicTrans fails to use the given context to disambiguate the gender of occupation terms and gives “male default” for most of the translations. Similarly, Microsoft and AWS Translate systems also show bias against women by translating most of the sentences into their ``male default" versions. Out of all the NMT models, Google Translate performs best at disambiguating gender from the given context. This shows that using such a set of sentences and extrinsic metrics, which take into account the gendered nature of the source sentence, is better at exposing the gender bias of an NMT system otherwise hidden by a metric such as TGBI.
    
    \subsection{Evaluation using WinoMT-Hindi}

        The results are shown in Table~\ref{tab:results-winomt}. Since \emph{Acc} i.e. Accuracy should be high while $\Delta_G$ and $\Delta_S$ values should be low, Google Translate outperforms other models as being the least gender-biased model. IndicTrans and AWS Translate are heavily biased toward a particular gender. These models have lower \emph{Acc} values (almost equal to the probability of a random guess, i.e. 50\%) and higher $\Delta_G$ values indicating that the $F_1$ score for the male class is very large in comparison to the $F_1$ score for female. 

        We also observe that $\Delta_S$ values are very low for all NMT systems. There are two potential reasons. First, it is observed that these HI-EN NMT systems strongly prefer masculine outputs irrespective of occupation stereotypes. Hence they give the ``masculine default'' in most cases leading to a similar performance on pro-stereotypical and anti-stereotypical sentences. Another reason can be the poor contextualisation of occupation stereotype. We rely on stereotype labels provided by original English occupation lists by \citet{zhao-etal-2018-gender} to divide the occupations into pro-stereotypical and anti-stereotypical sets. However, these lists were based on data from US Department of Labor. This might not contextualise well for Hindi. Culturally relevant occupation related statistics is required for creating these stereotype labels for different occupations in Hindi which was difficult to obtain in our case.
        
        However, WinoMT-Hindi provides a way to generalise and motivate the creation of such evaluation benchmarks for other languages.

 \section{Related Work}

Many works have focused on evaluating gender translation accuracy by creating various benchmarks. \textbf{WinoMT} benchmark by \citet{stanovsky-etal-2019-evaluating} is widely used for gender bias evaluation. It contains sentences from WinoBias \citep{zhao-etal-2018-gender} and Winogender \citep{rudinger-etal-2018-gender} coreference test sets in English. Without reference translations, it devises an automatic translation evaluation method for eight diverse target languages.

Other benchmarks include \textbf{MuST-SHE} \citep{bentivogli-etal-2020-gender}, \textbf{GeBioCorpus} \citep{costa-jussa-etal-2020-gebiotoolkit}, \textbf{MT-GenEval} \citep{currey-etal-2022-mt}, \textbf{GATE} \citep{rarrick2023gate} etc. MT-GenEval provides gender-balanced, counterfactual sentences in eight language pairs with English as the source.
Therefore, most of the benchmarks focus on English as the source language. 

Bias evaluation of NMT models on source languages other than English has mainly focused on the translation of gender-neutral sentences. \citet{cho-etal-2019-measuring} proposed \textit{TGBI} measure to evaluate gender bias in the translation of gender-neutral Korean sentences to English. \citet{ramesh-etal-2021-evaluating} used TGBI measure for Hindi-English machine translation. Our work emphasises on creation of gender unambiguous evaluation benchmarks for source languages other than English by accounting for gender inflections in the language to test the model's ability to find these gender-related cues. 
 
 \section{Conclusion and Future Work}
To conclude our study, we highlighted the need for contextualising NMT bias evaluation for non-English source languages, especially for languages that capture gender-related information in different forms. We demonstrated this using Hindi as a source language by creating evaluation benchmarks for HI-EN Machine Translation and comparing various state-of-the-art translation systems. In future, we plan to extend our evaluation to more languages and use natural sentences for evaluation without following a particular template. We are also looking forward to developing evaluation methods that are more inclusive of all gender identities.

\bibliography{anthology,custom}

\end{document}